\documentclass{article}
\usepackage{ijcai24}

\usepackage{times}
\usepackage{soul}
\usepackage{url}
\usepackage[hidelinks]{hyperref}
\usepackage[utf8]{inputenc}
\usepackage[small]{caption}
\usepackage{graphicx}
\usepackage{amsmath}
\usepackage{amsthm}
\usepackage{booktabs}
\usepackage{algorithm}
\usepackage{algorithmic}
\usepackage[switch]{lineno}
\usepackage{graphicx} 
\graphicspath{ {./figs/} }
\usepackage{xcolor}
\usepackage{multirow}

\title{A Survey of Robotic Language Grounding:\\Tradeoffs between Symbols and Embeddings}

\author{
Vanya Cohen$^1$
\and
Jason Xinyu Liu$^{2}$
\and
Raymond Mooney$^{1}$
\and
Stefanie Tellex$^{2,3}$
\And
David Watkins$^{3}$\\
\affiliations
$^1$UT Austin\\
$^2$Brown University\\
$^3$The AI Institute\\
\emails
\{vanya, mooney\}@cs.utexas.edu,
\{xliu53, stefie10\}@cs.brown.edu,\\
\{stefie10, dwatkins\}@theaiinstitute.com
}

\begin{document}
\maketitle

\begin{abstract}
With large language models, robots can understand language more flexibly and more capable than ever before.
This survey reviews and situates recent literature into a spectrum with two poles: 1) mapping between language and some manually defined formal representation of meaning, and 2) mapping between language and high-dimensional vector spaces that translate directly to low-level robot policy.
Using a formal representation allows the meaning of the language to be precisely represented, limits the size of the learning problem, and leads to a framework for interpretability and formal safety guarantees.
Methods that embed language and perceptual data into high-dimensional spaces avoid this manually specified symbolic structure and thus have the potential to be more general when fed enough data but require more data and computing to train.
We discuss the benefits and tradeoffs of each approach and finish by providing directions for future work that achieves the best of both worlds.
\end{abstract}

\section{Introduction}

Large language models (LLMs) have fueled a surge of interest in the problem of making robots understand natural language commands. Solving this problem requires mapping between words in language and actions or behaviors taken by the robot. \cite{harnad1990ground,harnad2007ground} defined the symbol grounding problem as constructing a mapping from symbols of a symbolic system or words in an utterance to sensorimotor substrates in the physical world.  Large language models (LLMs) semantically represent concepts without explicit higher-order symbols beyond the words in the text, leading many to try end-to-end approaches for robotic language understanding. Yet many recent works in robotic language understanding leverage large language models hand-in-hand with formal symbolic representations.  For example, Code as Policies~\cite{liang2022cap} generates Python code with predefined Python APIs. SayCan~\cite{ichter2022saycan} grounds natural language commands to predefined discrete skills implemented using a deep neural network. Other approaches take a more end-to-end approach, such as VIMA~\cite{jiang2023vima}, which learns a mapping from vision and language instructions to low-level robot actions such as joint states.

This survey paper evaluates work that grounds natural language to robot behavior.
We observe that these approaches can be situated on a spectrum ranging between two high-level approaches: mapping between language and a manually defined formal representation and mapping between language and high-dimensional vector spaces that translate directly to low-level robot policy.
We define the advantages and limitations of each approach.

Using a formal representation constrains the search space during training and inference and may require less training data.
It also provides a natural framework for strong interpretability and formal safety guarantees.
Well-defined model-checking tools exist to check whether the model of a system meets a given logical specification~\cite{baier2008principles}.
Formal methods can also synthesize correct-by-construction robot controllers given a logical specification and the system model and provide counterexamples to explain failure cases~\cite{kress2018synthesis}.  
However, a formal representation constrains the space of possible models that can be learned, limiting the system's ability to represent the meanings of what a person may say.
With the advent of large language models, mapping between human language and a formal language is much easier; the research questions then need to focus on what formal language to use, where it comes from, and how it connects to the physical world.
There are opportunities to more easily use existing representations such as the planning domain definition language (PDDL), linear temporal logic (LTL), or motion planners without collecting large training sets. Many approaches at the border, such as SayCan~\cite{ichter2022saycan}, map language to a manually specified vocabulary of robot skills but give their formal language relatively little attention despite it playing a critical role in the system.

\begin{figure*}[ht]
  \centering
  \includegraphics[width=\linewidth]{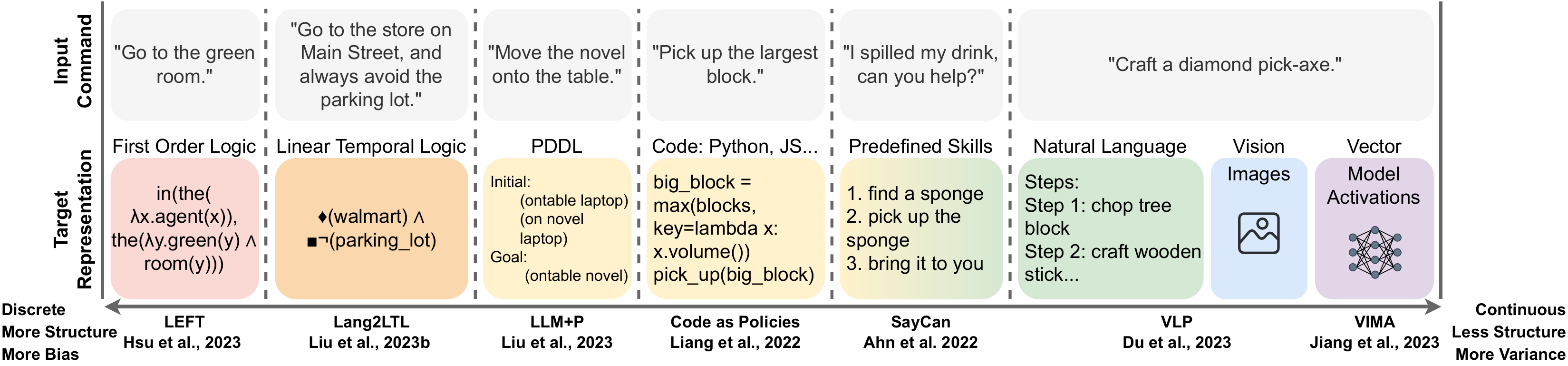}
  \caption{Approaches to representing natural language for robotics fall along a spectrum from more symbol-like representations to more continuous embedding-like representations. However, most approaches use a mixture of both. SayCan uses a fixed ontology of predefined skills but implements these as neural value functions conditioned on language.}
  \label{fig:spectrum}
\end{figure*}

End-to-end approaches may require more data to train but are more flexible in representing a user's intended meaning and translating it to robot behavior.
\cite{ng2001discriminative} observed that structured generative models perform better with less training data.
In contrast, discriminative models with more parameters can perform better when given lots of data because they make fewer assumptions about the structure of the learned model.
Similarly, end-to-end neural embedding approaches require more data but can generalize better than formal methods because they place fewer constraints on the learned model.
True ``pixels to torque'' approaches~\cite{wahlstrom2015pixels} learn to produce motor torques directly from sensor input; however, many end-to-end approaches use intermediate outputs such as end-effector poses or joint states.
The challenge with end-to-end approaches is acquiring enough training data, processing this data, generalizing outside the training set, chaining policies together to produce long-term behaviors, explaining robot behaviors, and providing safety guarantees.

We situate the robotic language grounding works on a spectrum in Figure~\ref{fig:spectrum} and describe more formal work in Section~\ref{sec:formal} and more end-to-end work in Section~\ref{sec:p2t}.
More formal work uses discrete and structured representations that introduce bias into the learning.
We begin with the most structured, abstract representations and successively review less structured, lower-level, continuous representations.
Many of the implementations surveyed combine aspects of formal and end-to-end representations.
In Section~\ref{sec:discussion}, we survey output representations for both approaches, review available datasets, and discuss the scope of natural language commands understood in theory and practice by different approaches.
We conclude by identifying key open problems and recommending future research directions.

This new review paper focuses on the transformative role LLMs have played in this space. Other relevant reviews are:
\begin{itemize}
    \item \cite{tellex2020robots} is a review that surveys robot and language grounding work predating LLMs' emergence.
    \item \cite{zhang2023large} reviews large language models more broadly used in human-robot interaction, including question answering, social robotics, and instruction following.
    \item \cite{zeng2023large} reviews LLMs applied to robotics broadly, including related technologies, but does not focus on the spectrum from formal methods to high-dimensional vectors as this paper does.
    \item \cite{wang2024large} reviews applications of LLMs to robotics but does not situate the work on a spectrum and focuses on a broader set of tasks than command understanding.
\end{itemize}
Our work, in contrast, focuses specifically on the problem of command understanding, situating work along a spectrum based on formal methods.

\setlength{\tabcolsep}{3pt}
\begin{table*}[ht]
\scriptsize
\begin{tabular}{@{}llllll@{}}
\toprule
                                \multirow{3}{*}{}  & Formal & Intermediate & Final System & Executing Output on Robots & Domains \\ & vs. Intermediate & Grounding & Output \\
                                & vs. End-to-End \\
\midrule
LEFT~\cite{hsu2023whats}        & Formal        & N/A & FOL formula       & Low-level controllers & Pick-and-Place (Real) \\
Lang2LTL~\cite{liu2023lang2ltl}        & Formal        & N/A & LTL formula       & LTL-MDP planner, Low-level controllers & Navigation (Real) \\
\multirow{2}{*}{\cite{xie2023translating}}        & Formal        & N/A & PDDL goal       & Symbolic planner, Low-level controllers & Navigation, \\
& & & & & Mobile manipulation (Sim) \\
LLM+P~\cite{liu2023llm+p}        & Formal        & N/A & PDDL       & Symbolic planner, Low-level controllers & Manipulation (Real) \\
SayCan~\cite{ichter2022saycan}         & Formal        & N/A & predefined skill   & Low-level controllers & Mobile manipulation (Real) \\
\multirow{2}{*}{Code as Policies~\cite{liang2022cap}} & Formal        & N/A & Python code          & Python interpreter, Low-level controllers & Pick-and-place, Mobile manipulation \\ & & & & &  Drawing (Real) \\
ProgPrompt~\cite{singh2023progprompt} & Formal        & N/A & Python code          & Python interpreter, Low-level controllers & Mobile manipulation (Sim) \\ & & & & & Manipulation (Real)  \\
Voyager \cite{wang2023voyager}         & Formal        &  N/A & Javascript Code      & Javascript interpreter, Low-level controllers & Minecraft survival tasks (Sim) \\
ITP~\cite{li2023interactive} & Intermediate        & High-level action & Python code          & Python interpreter, Low-level controllers & Manipulation (Real) \\ & & descriptions & & &  \\
UniSim~\cite{yang2023learning}         & Intermediate    & Image subgoals & predefined skill & Low-level controllers & Manipulation (Real) \\
SuSIE~\cite{black2023zeroshot}         & Intermediate    & Image subgoals &  End-effector pose & Low-level controllers & Manipulation (Real) \\
VLP~\cite{du2023video}                 & Intermediate    & Image subgoals &  End-effector pose & Low-level controllers & Manipulation (Real) \\
PaLM-E~\cite{driess2023palm}           & End-to-End    &  N/A & predefined skill   & Low-level controllers & Pick-and-place, Manipulation (Real) \\
VIMA~\cite{jiang2023vima}                       & End-to-End & N/A & predefined skill & Low-level controllers & Manipulation (Sim) \\
PerAct~\cite{shridhar2023peract}       & End-to-End    & N/A & End-effector pose   & Low-level controllers & Manipulation (Real) \\
\multirow{2}{*}{RT-1~\cite{brohan2023rt1}} & End-to-End & N/A & End-effector pose   & Low-level controllers & Manipulation (Sim, Real) \\
& & & & &  Mobile Manipulation (Real) \\
RT-2~\cite{zitkovich2023rt2} & End-to-End & N/A & End-effector pose   & Low-level controllers & Manipulation (Real) \\
RT-X~\cite{padalkar2024rtx}                    & End-to-End    & N/A & End-effector pose   & Low-level controllers & Manipulation (Real) \\
\bottomrule
\end{tabular}
\caption{Method Comparison}
\label{tab:methods}
\end{table*}

\section{Mapping from Natural Language to a Formal Representation}\label{sec:formal}





Works closer to the formal end of the spectrum map natural language commands from humans to a manually defined formal representation, e.g., temporal logic, planning domain definition language (PDDL), computer code, or some predefined skills.
The symbols in the formal representation are then grounded to robot percepts and control by predefined detectors and controllers, respectively.
Unlike machine translation of natural languages, where vast training data is available online, translating natural language to logic often lacks labeled pairs of natural language commands and logic formulas.
Most recent works leverage few-shot learning or fine-tuning of large language models (LLMs) for various parts of the language grounding system to address the lack of training data.

Many formal representations used to ground natural language are Turing-complete and thus can be translated from one to another.
However, depending on the language used, this translation may be direct or indirect and require significantly longer or more complex expressions.
For example, linear temporal logic (LTL) can naturally represent English sentences such as ``Avoid the red room'' with a short, direct expression.
Python can represent the same command by defining an ``avoid'' function but may need a significantly longer program if the function is not provided.
Thus, in our review, we order these representations from those with more structure and bias to those with less structure and less bias, as typically used, shown in Figure~\ref{fig:spectrum}.

More structured methods often map directly to certain natural language commands and express the goal or constraint directly rather than imperatively how to achieve it.
Goal-based representations specify what state the world should be in but not what actions the robot should take to attain that state.
In contrast, action-based representations specify a sequence of actions but not necessarily the goal or results of those actions.
A goal-based representation for ``avoid the red room'' might be a logical formula such as $\neg red\_room$ while an action-based expression might be $North; North; West; West; North; North$ (depending on the specific geometry of the environment).  
We order our review from high-level, abstract representations to low-level, concrete, fine-grained representations.

\subsection{Logics}
Logics are mathematically precise goal-based representations that specify robotic goals and provide guarantees for robot behaviors.   
Temporal logics can concisely represent long-horizon, temporally extended tasks.
\cite{kress2018synthesis} surveyed the uses of several temporal logics as task specifications for the formal synthesis of robot controllers.
We consider logical expressions at the most formal end of the spectrum because they map to goals, describing abstractly the state of the world corresponding to the language, leaving the plan to achieve this state of the world to other modules. 

To train their language grounding system on diverse natural language commands, \cite{pan2023data} used LLMs to paraphrase structured English commands generated from algorithmically produced LTL formulas.
\cite{wang2021learning} and \cite{patel2020grounding} trained a semantic parser to map language commands to LTL formulas using weak supervision of execution trajectories without any LTL annotations.
Lang2LTL~\cite{liu2023lang2ltl} is a modular system that uses LLMs to ground navigation commands to linear temporal logic (LTL) formulas and their propositions to physical landmarks in a given semantic map.
The same system solved navigational tasks in indoor and outdoor environments without retraining on language data by harnessing pre-trained LLMs.
Similarly, \cite{hsiung2022generalizing} first translated commands to lifted LTL formulas and then grounded them to specific domains for better generalization.
Other approaches~\cite{fuggitti2023nl2ltl,chen2023nl2tl} also used LLMs to translate natural language commands to logical representations but did not ground the formulas to a robot domain.

AutoTAMP~\cite{chen2024autotamp} uses LLMs to translate task and state descriptions to signal temporal logic (STL) formulas~\cite{maler2004monitoring} and correct syntax errors if detected.
It then uses an STL planner to generate trajectories.
By using an intermediate formal task specification, AutoTAMP outperforms LLM planners on tasks with geometric and temporal constraints in 2D domains.

Recent work also leveraged LLMs for grounding natural language commands to first-order logic (FOL) formulas.
LEFT~\cite{hsu2023whats} used an LLM to translate natural language queries to FOL programs, which a differentiable FOL executor executed.
At the same time, a domain-specific grounding model grounded the symbols of the FOL program to various input modalities, e.g., 2D images and point clouds.
The advantage of using first-order logic over LTL is capturing commands that use quantifiers and predicates for generalization.
On the other hand, LTL provides a natural way to concisely represent temporally extended commands.
Different logic representations can be formally composed to form a new, more expressive logic representation.
These representations can enable direct mapping between abstract concepts in a language such as ``avoid'' and a precise, formal logical expression that guarantees task satisfaction.




\subsection{Planning Domain Definition Language (PDDL)}
The Planning Domain Definition Language (PDDL)~\cite{mcdermott1998pddl,fox2003pddl2,edelkamp2004pddl2,kovacs2011pddl3} is a structured representation that defines a planning problem.
It consists of a domain, which defines objects, predicates, and actions that govern the world's rules, and a problem, a grounded problem instance with an initial state and a desired goal state.
A symbolic planner takes as input the PDDL domain and problem and then outputs a plan~\cite{helmert2006fast}, i.e., a sequence of actions, to reach the goal state from the initial state.
In this sense, it is a goal-based representation, but because it encodes actions and their effects, it can also be used imperatively.
Recent works used LLMs to translate natural language descriptions of the world and the task to a PDDL representation of the planning problem, which a symbolic planner can then use.
Compared to logical representations, PDDL provides a language of predicate states and transitions that allow goals to be translated into lower-level actions and skills but requires a full domain specification.
People have defined PDDL domains for a wide variety of real-world applications.  \cite{konidaris2018skills} showed how a robot can learn symbols for low-level skills that are both necessary and sufficient to enable planning in PDDL.  

\cite{xie2023translating} prompted LLMs with a PDDL domain description, an initial state, and examples to translate a natural language goal description to a PDDL goal specification.
Results in simulation showed that LLMs could translate unambiguous goal descriptions and fill in missing details for under-specified goals but struggled with numerical and spatial reasoning.
\cite{collins2022llm} also showed that translating natural language goal descriptions to PDDL goals and then solving them using a symbolic planner outperformed directly using an LLM as a planner.

\cite{guan2023leveraging} and \cite{liu2023llm+p} prompted an LLM to translate a natural language description of a problem into a complete PDDL problem definition, which was then fed into a symbolic planner together with a PDDL domain definition to find an optimal plan.
These approaches outperform LLM planners and provide strong correctness guarantees from using a symbolic planner.
Recent work has also created benchmarks for evaluating LLM's ability to make plans concerning PDDL and other AI baselines~\cite{valmeekam2023planbench,valmeekam2023planning}, finding that LLMs achieve low success rates across domains in isolation.
Still, they can improve the search process for underlying sound planners, leading to higher performance.

Unlike other works that use LLMs for AI planning, \cite{silver2024generalized} developed a generalized planner to synthesize Python programs to solve novel tasks by prompting LLMs with a domain specification and a few training tasks emphasizing satisfaction and efficiency.
Their system also automatically detects planning errors and then re-prompts the LLM with feedback.

\subsection{Code}
Code is a very flexible form of formal representation that can be used in a goal-based or an action-based manner.
Indeed, a Python program can be generated that directly outputs motor torques for robots or implements an arbitrary deep-learned function.
A formal language is typically specified as a subset of the language at hand using a manually defined programming API consisting of functions and their arguments analogous to robot skills.
Skills or functions are not simply linearly called but can be embedded in more complex logic, like conditionals and loops.

Code as Policies~\cite{liang2022cap} prompted an LLM with import statements, example code, and code comments that describe the desired policy to generate Python code executable directly on the robot.
It solved manipulation and mobile manipulation tasks with API calls to Python libraries and predefined perception and control modules.
ProgPrompt~\cite{singh2023progprompt} applied a similar approach with additional assertion statements to recover from errors when reliable state tracking is available.
\cite{varley2024embodied} developed a modular bi-arm system that employs an LLM to map natural language instructions to high-level API calls to manipulation skills powered by VLMs and a control module to solve three tabletop bimanual manipulation tasks.
Their experiment results highlighted the benefits of modularity for ensuring safety, interpreting failures, and identifying modules to improve.
Socratic Models~\cite{zeng2023socratic} also demonstrated the code generation capabilities of LLMs for simulated pick-and-place tasks, provided with pre-trained perception modules and robot policies.
ITP~\cite{li2023interactive} used LLMs to generate high-level action sequences and then translated the action descriptions to predefined function calls using APIs for an object detector and robot policies to solve tabletop manipulation tasks.
By storing completed high-level actions, ITP can replan given new commands at any step during the execution.
Voyager~\cite{wang2023voyager} extended the code generation capabilities of LLMs to build a lifelong learning agent in Minecraft that continuously explores and learns a skill library of executable code.

\subsection{Predefined Skills}
Recent methods have used LLMs as a planner to map language commands to a sequence of predefined skills.
Skills can be learned from data or manually specified.
These predefined skills are a formal representation since they are discrete and manually specified.
(Even if skills are learned, they are frequently learned from human-provided demonstrations, which provide the discrete structure for the skills.)
They are also action-based representations specifying actions to take rather than goals or resultant states.
For example, a robot might be provided with skills such as ``pick up,'' ``put down,'' ``drive to refrigerator,'' ``drive to microwave,'' and ``clean table.''
Then, the challenge is to map a natural language instruction such as ``Get me the apple'' and ``Clean the table with the sponge'' to a sequence of skills.
\cite{huang2022language} iteratively prompted an LLM to decompose a high-level task specification in natural language to a sequence of action descriptions.
They used sentence similarities to map the proposed action descriptions to actions available in a simulated environment where their corresponding predefined low-level controllers can be executed.
SayCan~\cite{ichter2022saycan} iteratively prompted an LLM to sequence pre-trained skills described by predefined verb phrases based on their probabilities of success from the current state to solve mobile manipulation tasks specified by natural language on a physical robot.
Their approach relies on both formal representations and high-dimensional, end-to-end differentiable representations.
Despite using a fixed ontology of skills, these are implemented using a deep approach.
SayCan learns these skills using a multi-task value function conditioned on language.
CAPE~\cite{raman2024planning} also used an LLM planner to sequence predefined skills. When a plan does not meet the precondition of some skill, it re-prompted the LLM with corrective feedback.
In addition to using an LLM to ground language to predefined skills, Inner Monologue~\cite{huang2022inner} also enabled language feedback based on pre-trained perception models for describing the scene and detecting successful execution of skills.
In these methods, the specific skills are often given relatively little attention, yet having the right set of skills at the right level of granularity is critical to success. We categorize this method on the formal side of the spectrum because the skills impose significant structure.  That said, it imposes less structure than goal-based methods such as logical representations, leaving the LLM to keep track of higher-level constructs such as constraints, sequencing, and conditionals.




\section{Mapping from Natural Language to High-Dimensional Vectors to Actions} \label{sec:p2t}

On the other end of the spectrum are approaches that map natural language instructions to high-dimensional, non-formal representations.
These deep approaches are largely defined by data and learning, whereas formal, symbolic representations are largely human-crafted and invoke manually provided structure.
The large models that enable deep and many symbolic approaches are created by self-supervised pretraining on large datasets scraped from the web.
These models thereby learn a generative representation of images, text, and other modalities that capture background and task knowledge that is useful for robotic applications, for example, as in ~\cite{driess2023palm}.
By learning to model a large and varied distribution, the models also learn to perform tasks useful for robotic language grounding, including translation, semantic parsing, and broadly useful ones like in-context learning~\cite{brown2020gpt3}.

Pretrained models can represent high-level and low-level semantics, so they can represent high-level instructions and low-level actions.
This semantic understanding can extend across modalities while retaining these task-learning abilities, forming combined representations of language and vision useful to robotics problems.
Beyond their representation capacity, a unified interface for training and inference makes these methods particularly powerful.
Multiple models across modalities can be connected and trained end-to-end through gradient descent, and pretrained representations can be improved through increases in model and data scale ~\cite{brown2020gpt3,henighan2020scaling,hoffmann2022scaling,wei2022emergent}.
Provided with a method for generating and standardizing data, a single pipeline with minimal human intervention can be used for continued improvement and deployment.
One limitation of deep methods is that they require abundant data, which limits their applicability to domains where data is easy to collect.
However, large-scale pretraining can significantly reduce the amount of task-specific training data required~\cite{brown2020gpt3}.

In robotic language grounding, deep learning approaches map input language commands to one of the following representation types: actions represented as low-level controls (policies) or high-level goals represented as natural language, images, or neural network activations. These high-level goals are used to condition low-level actions from planners or policies to drive robot actions. At the lowest level, the approaches map language to sequences of joint states, end-effector poses, or motor torques. High levels of abstraction are easier for humans to interpret versus low-level joint states, making it difficult to discern the outcome of the action. We order from high-level to low-level to indicate the levels of abstraction as they have evolved in the field and to create a spectrum for interpretability.

\subsection{Image and Language Subgoals}
These methods use images and natural language to express subgoals. \cite{black2023zeroshot} perform pick-and-place tasks on a robot by alternating between generating images representing sub-goals using a text-guided image-editing model and executing a low-level policy conditioned on the goal images.
They map language to subgoals represented as images and then map these images to end-effector positions.
VLP~\cite{du2023video} transforms the input image and command into a sequence of language and image subgoals and then generates plan rollouts with tree search using a language-conditioned video generation model.
It was deployed on a robot to perform tabletop arrangement tasks.

UniSim~\cite{yang2023learning} trains a generative video model to predict the outcome of both high and low-level actions.
Using this model, the authors train high-level planners and low-level policies in simulation and demonstrate zero-shot transfer to real-world autonomous driving tasks.
Language and images are mapped to high-level skills and low-level controls, e.g., ``move the gripper to (x, y)''. 
GAIA-1~\cite{hu2023gaia1} similarly leverages real-world driving data to develop a world model that generates life-like driving behavior by predicting outcomes in the video space.
Coarse symbols, such as words in English, are not as rich of a representation as images.
Incorporating multimodality increases the generalizability and expressiveness of the systems' states. This allows for directly representing the visual world state without introducing an intermediate representation like natural language, scene graphs, or maps. The downside to using images is that it requires more pretraining and fine-tuning data to learn this multimodal representation.

\subsection{End-Effector and Joint-State Goals}
These approaches use low-level skills and parameterize actions using joint states or end-effector position, in contrast to high-level skills like ``pick up'' that we classify on the formal side of the spectrum. Often, they are not learned because they are sufficiently low-level, and the built-in controllers on the robot are sufficient.  For example, the RT-* family of models~\cite{brohan2023rt1,zitkovich2023rt2,padalkar2024rtx} use an action space that specifies the arm, base, and end-of-episode token. For the arm, they use the end-effector pose of the gripper (vs. lower-level joint states or joint torques) and rely on inverse kinematics to find joint states to move to the end-effector pose.  

RT-1, RT-2, and RT-X~\cite{brohan2023rt1,zitkovich2023rt2,padalkar2024rtx} perform various language-conditioned tasks on 22 distinct robotic platforms.
They collected a large and diverse dataset of demonstrations and trained a large multimodal transformer that maps language and state observations to low-level discrete end-effector controls. These methods attempt to demonstrate positive transfer, the transfer of knowledge between tasks that allows a system to perform better at both tasks simultaneously.
The Open-X Embodiment dataset~\cite{padalkar2024rtx} release came with baselines showing this phenomenon, whereas GATO~\cite{reed2022gato} did not.
Open-X Embodiment targeted end-effector positions and used models from 35M to 55B parameters, whereas GATO targeted joint states directly and used 1.2B parameters. 
Octo~\cite{octo_2023} also shows better performance across tasks by leveraging the Open-X Embodiment dataset while simultaneously outputting end-effector positions and joint-state control.
PaLM-E~\cite{driess2023palm} is an image and text multimodal model finetuned end-to-end for various multimodal reasoning tasks, including robot manipulation.
This model maps visual-language input describing a scene to a limited vocabulary of actions, which are then turned into robot actions. VIMA~\cite{jiang2023vima} learns a multimodal model that maps tasks specified with images and language to high-level actions parameterized by low-level end-effector controls to perform tabletop arrangement tasks on a simulated robot.
One consequence of this level of abstraction is that what is learned is often specific to a robotic platform and cannot be generalized from one environment and robot to another without more data.
Generating data automatically in simulation allowed it to overcome limitations on the data required to learn a meaningful representation.
Future work is focused on increasing the size and diversity of the datasets to enable more generalization.

\vspace{-0.3mm}

Recent works in foundation models have shown tremendous improvement in generalization by leveraging large amounts of data.
These methods often use symbols in the form of vector-quantized image patches to reduce the number of tokens required~\cite{yan2023temporally}. 
ALOHA~\cite{zhao2023learning} seeks to address the limited data for training these systems by utilizing accessible hardware for many research institutions. They then use an end-to-end foundation model that targets joint-state control of robots to execute policies on the bimanual robot platform. 
\cite{li2023vision} addresses domain limitations by leveraging vision and text models that are larger and trained on more data.
\cite{shridhar2023peract} develop a language-conditioned behavior-cloning model that encodes the inputs of a natural language command and a voxel grid to predict a collision-free pose that a motion planner can execute.
Similarly, VPT~\cite{baker2022video} utilizes real demonstrations of users playing Minecraft to train better policies that target a discrete set of actions for the agent to follow.
They expand their dataset by training an inverse dynamics model to label data from YouTube.
Joint-state and end-effector goals are useful for simultaneously targeting multiple robotic platforms but can lack the interpretability of their output.

\setlength{\tabcolsep}{3pt}
\begin{table*}[ht]
\tiny
\centering
\begin{tabular}{@{}llp{8cm}p{5cm}@{}}
\toprule
                                \multirow{2}{*}{}  Command Types & Domains & Command Examples & Implementation Examples \\
\midrule
\multirow{3}{*}{Temporal} 
& Navigation & ``Move back and forth between the table and the countertop twice.'' & \cite{liang2022cap} \\
& & ``Go to the store on Main Street, but only after visiting the bank'' & \cite{liu2023lang2ltl} \\
& & ``Remain on the first floor and navigate to the red room .'' & \cite{pan2023data} \\
& Manipulation & ``Scan for blocks and insert any found into the bin.''  & \cite{gopalan2018sequence} \\
& & ``Create a stack that contains two blocks.'' & \cite{xie2023translating} \\
& Mobile Manipulation & ``Get Glass of Milk'' & \cite{huang2022language} \\
& & ``Always take the pear and go to the tree and stay there.'' & \cite{wang2021learning} \\
& & ``I spilled my coke. Can you bring me something to clean it up?'' & \cite{ichter2022saycan} \\
& & ``Can you bring me the drink from the table?'' & \cite{huang2022inner} \\
& & ``Bring the plate to the kitchen'' & \cite{hsiung2022generalizing} \\
& & ``Move KeyChain1 to the box with more books'' & \cite{xie2023translating} \\

\midrule

\multirow{2}{*}{Spatiotemporal} 
& Navigation & ``Go to Landmark A then Landmark B, in addition, avoid Landmark C;'' ``Move back and forth between Object A in Room B and Object C in Room D'' & \cite{liu2023lang2ltl,pan2023data} \\
& Manipulation & ``Put the red block to the left of the rightmost bowl'' & \cite{liang2022cap} \\
& & ``Pack the ring into the brown box'' & \cite{hsu2023whats} \\
& & ``Sort fruits on the plate and bottles in the box'' & \cite{singh2023progprompt} \\
& & ``Move all the blocks to different corners.'' & \cite{zeng2023socratic} \\
& & ``May I have a cup of milk with taro?'' & \cite{li2023interactive} \\
& & ``Stack Objects X, Y, and Z'' ``Push Object X to Object Y'' & \cite{black2023zeroshot,brohan2023rt1,driess2023palm,black2023zeroshot,yang2023learning,jiang2023vima,liang2022cap,du2023video,zitkovich2023rt2,padalkar2024rtx} \\
& Mobile Manipulation & ``Take the Coca-Cola can from the desk and put it in the middle of fruits on the table'' & \cite{liang2022cap} \\
& & ``Microwave salmon'' & \cite{singh2023progprompt} \\
& & ``Bring Object A to Location B;'' ``Open the door and proceed to the kitchen'' & \cite{ichter2022saycan,driess2023palm,huang2022inner,hsiung2022generalizing}  \\
& & ``Build a house out of dirt blocks'' & \cite{wang2023voyager} \\
& & ``Put away groceries in the pantry'' & \cite{huang2022language,li2022pre} \\
\bottomrule
\end{tabular}
\caption{Situated Natural Language Commands}
\label{tab:commands}
\end{table*}

\section{Discussion and Future Directions}\label{sec:discussion}

Each end of the spectrum has tradeoffs and advantages, and the two approaches are complementary in many ways.
As shown, many recent works combine aspects of symbolic and end-to-end methods.
A number of open problems and exciting future research directions are to leverage the best of both worlds.

\subsection{Representation}
Table~\ref{tab:methods} summarizes the grounding representations used by various systems we review and how they execute their output on a robot in various simulated and real-world domains.
On the formal side of the spectrum, system output ranges from more structured logic formulas and code to less structured predefined skills.
On the deep side of the spectrum, methods ground language input to high-dimensional representations of joint states or end-effector poses.
Some deep approaches first ground language to an intermediate representation of image or language subgoals, then low-level controls.
Formal approaches to executing the system output on a robot use a planner or a code interpreter together with low-level controllers.
In contrast, deep approaches use low-level controllers to produce desired robot motion.
A key takeaway is using intermediate representations at different system levels, which enables the use of ``off-the-self'' robot modules such as SLAM and object detectors.
When using these modules, there is a crisp abstraction barrier between the learning model and the robot, limiting what needs to be learned and the flexibility of the learned system.
Automatically learning an extendable set of symbols and grounding those symbols~\cite{gopalan2020symbol} can address this limitation of formal approaches.
Another key research question for formal representations is what formal language to use.
Yet a formal language that is powerful and flexible enough to capture all English or other natural languages is unknown.
Approaches such as RLANG~\cite{rodriguez2023rlang} expand the scope of what language can talk about (from goals and actions to observations, states, and transition functions).
Integrating these formal, manually specified languages with open-class learned models is critical, for example, jointly learning skills and symbols~\cite{konidaris2018skills,gopalan2020simultaneously}.
This survey only focuses on natural language, yet using multimodal input, e.g., text, audio, RGB-D images, videos, and joint trajectories, etc., to complement language~\cite{reed2022gato} in solving robotic tasks is also a promising future research direction. Current state-of-the-art approaches combine multiple modalities in a single neural architecture capable of running in real-time, including speech, image, and text~\cite{reid2024gemini}. This promises to enable more audio interactivity between humans and robots.

\subsection{Situated Natural Language Commands}
Table ~\ref{tab:commands} reviews various domains and natural language commands used to evaluate each system we review.
Domains vary from tabletop manipulation tasks, usually focusing on pick-and-place, to mobile manipulation, usually involving chaining pick-and-place actions and navigation, in both simulation and the physical world.
Table~\ref{tab:commands} shows examples of natural language commands that specify temporally extended tasks and spatial relations among objects in various domains.
Natural language was created based on an assumption about human reaction time. The methods described in this work have control loops in the space of 1-3 Hz, which cannot close the loop fast enough to react to sentences such as ``move to the left'' followed by ``okay, stop.'' For real-time robotics applications, faster language processing methods need to be created. These could be through the advent of faster neural network hardware or methods that learn hierarchical abstraction at different frame rates to support the cross-cutting ability of language to talk about any part of the robot system.

\subsection{Datasets}
Both approaches require datasets for learning and evaluation.
Formal methods generally require parallel datasets that map natural language to structures in the formal representation or at least the ability to test if a formal representation is correct.
These parallel corpora can be expensive at scale, but new LLM methods enable good performance even with much smaller datasets.  
A common approach is to show a trajectory of robot behavior generated from an underlying formal representation and then ask annotators to describe the trajectory in language~\cite{gopalan2018sequence,patel2020grounding}.
This approach can enable untrained annotators to provide parallel data.
Still, it often leads to ambiguous situations because the trajectory does not overtly show the constraints present in the underlying representation.
Thus, it remains an open problem to collect diverse and unambiguous language commands and their corresponding labels.

To compensate for this problem, approaches, especially less formal approaches, try to learn from large unannotated datasets or from datasets obtained from simulation.
LLMs can also enable robust learning to map from natural language to formal representation using few-shot learning or fine-tuning~\cite{liu2023lang2ltl,hsu2023whats,liu2023llm+p,liang2022cap,ichter2022saycan}.
Methods that use less constrained representations, such as VIMA~\cite{jiang2023vima}, collect data in simulation across many scenarios and eschew a formal representation that requires annotation of any kind, instead directly train from joint trajectories, images, and video.

\subsection{Generalization and Bias}

Models that map language to a structured representation can exploit the regularities of that structure to train from smaller datasets and can generalize by porting the formal representation to different domains.  Depending on the domain, this generalization can be significant by relying on state-of-the-art robot algorithms such as SLAM and motion planning rather than trying to learn everything end-to-end.  This ability is possible because of the strict, modular compositionality introduced by the formal model.        

On the other hand, end-to-end models have the potential to be applied in domains given large enough training datasets and computational resources.  Looking at the power of LLMs, one predicts that a multimodal dataset of video, other robot sensor data, and joint states may be able to generalize as flexibly and powerfully as an LLM does with language.  However, training this model requires significantly higher dimensions than language data and potentially much larger dataset sizes.  However, if provided with general-purpose robot data, foundation models have the potential to understand language in a very general way, analogous to the success of LLMs. 
Key research challenges to address this problem include acquiring semantically diverse datasets covering desired robot behaviors, developing sample-efficient model architectures and training algorithms for multimodal datasets many times larger than those used to train LLMs, and translating learned models to robot action.

Many existing works that use foundation models also leverage symbols but do not speak to the symbols' role in building practical systems.
We observe that in many recent RoboNLP papers including ~\cite{ichter2022saycan,jiang2023vima,brohan2023rt1,zitkovich2023rt2,padalkar2024rtx} all leverage symbolic representations, either in the form of discretized action spaces or predefined skills.
We would like to see more authors acknowledge that their work is made better because of the symbolic representation their model uses or propose how other symbolic representations can improve the model.

In the future, we expect models to better represent the world through these finer-grained symbolic representations.
After a certain point, combining text, audio, color images, depth images, point clouds, etc., becomes functionally continuous.
While the POMDP and PDDL methods are theoretically sound, in practice, they lack representational capacity outside of their domain due to the brittleness of out-of-domain representation.
We believe that future successful methods will leverage deep learning methods with smarter representations of the input and output data via transformations of the intermediate representations.
This includes representing color images as point clouds, maps, or meshes as derivative intermediates.
We can then augment these with state representations of the environment as text or other labels.
There is an eventuality where the bitter lesson~\cite{sutton2019bitter} takes over, and throwing even more parameters and data at the problem will win out, but that is predicated on an evergrowing data supply.
It remains to be seen where additional data will come from and if methods that leverage only partial observations from one modality can synergize multiple modalities simultaneously.

\subsection{Limitations of Natural Langauge}
As robot operators, we want to be able to express the state of our robotic system in a way that is auditable by us and useful to the robot in accomplishing tasks.
Many of the methods we discussed also focus on English as the representation of natural language encoded in LLMs and multimodal models.
While English works well for much of the world, it does not precisely describe objects' physical locality.
Saying that an apple is to the left of an orange can put that apple in various positions with just a language description alone.
While adding geometric priors could enhance the specificity of the language, we should, as a field, look to building benchmarks to evaluate the physical plausibility of these language models as they operate in different languages~\cite{whorf1956language,lucy1992language}.
Robot morphology and human morphology are very different. It is likely that expressing tasks as being completed by one's arm, such as ``clean the dishes with your left hand'' may be completely moot to a robot that uses soft-body mechanics to interact with the world.
Human language evolved in a biological domain that necessitates a more survival vocabulary.
These actions are not necessarily helpful for describing robot actions and could contribute to skewing the success rates of robots in accomplishing tasks.

\subsection{Safety and Interpretability}
Interpretable and explainable robotic systems provide transparency in decision-making and gain trust in human-robot interaction~\cite{gunning2019darpa,anjomshoae2019explainable,silva2023explainable}. 
Moreover, verifiable safe operation is essential for deployments that satisfy worldwide standards such as ISO 61508~\cite{iso61508}, which defines standards for safely deploying robots in industrial factory environments worldwide.
These standards require that robotic systems be mathematically proven to have a failure rate lower than $10^{-5}$ dangerous failures per hour.
Robots should also provide feedback to unsatisfiable task specifications~\cite{raman2013sorry} or explain their actions when execution fails~\cite{das2021explainable}.
Safety can be classified into semantic and kinematic/physical safety~\cite{varley2024embodied}.
Examples of semantic safety are never entering the nursery and always transporting a cup full of coffee in an upright orientation.
Examples of kinematic safety are avoiding collisions with humans and objects and avoiding reaching joint/velocity/torque limits.
Formal approaches using LLMs mostly focus on enforcing semantic safety.
Many works on formal methods and safe control do not use LLMs and encode kinematic safety in trajectory optimization~\cite{dawson2023safe}.
For example, linear temporal logic (LTL) has been used extensively to develop provably safe controllers~\cite{lignos2015provably,chinchali2012provably}.
Given a logical specification and the system model, formal methods can synthesize correct-by-construction robot controllers or provide counterexamples to explain when the task is unsatisfiable with respect to how well the model captures reality~\cite{kress2018synthesis}.
A key challenge for any safety framework of this kind is grounding predicates in real-world noisy and partial perceptual data.

Existing deep learning approaches do not have a clear pathway for achieving this level of interpretability and safety.
If we look at human behavior, defining a set of safety guidelines to assess adherence has been challenging.
While we can continue to apply formal representations to these models to varying levels of success, there remains a need for a system that, at a meta-level, operates on symbols but whose abilities are as general and flexible as humanity's. For example, one of our recent papers uses a combined approach with LLMs and a formal representation, showing that we can exploit the generalization ability of LLMs with the formal safety guarantees provided by LTL to create a safe yet robust and flexible system for following commands~\cite{yang2024plug}. Yet this approach only scratches the surface.
Much more needs to be done to integrate these systems, especially to study the interpretability and safety guarantees inherent in perceptual systems and provide formal guarantees and bounds about the behavior of deep networks.

\section{Conclusion}
Our review characterizes the literature in robotic language grounding along a spectrum from using more formal, biased, discrete representations to less formal, less biased, higher-dimensional continuous representations.
There are benefits and tradeoffs to each approach.
More formal methods induce structure that can limit the size of the learning problem and provide interpretability and formal safety guarantees.
However, they also constrain the output space, limiting the flexibility and expressive power of what can be learned.
Less formal methods impose fewer constraints but require more data and possibly more structured neural networks to be learned. 
Methods such as SayCan~\cite{ichter2022saycan}, traditionally considered less structured, use a formal representation of the robot's skills.
Seen in this context, it is clear that a limitation for all methods is the lack of physical capability of existing robots: a key area of future work is to enable robots to perform a larger variety of tasks in a larger variety of environments. Hence, they can physically perform the tasks people ask them to do.

\section*{Acknowledgments}
The work done by Vanya Cohen and Raymond Mooney is supported by the Defense Advanced Research Projects Agency (DARPA) under Contract No. HR001122C0007.
The work done by Jason Xinyu Liu and Stefanie Tellex is supported by the Office of Naval Research (ONR) under grant number N00014-22-1-2592 and with support from Amazon Robotics.
Any opinions, findings, conclusions, or recommendations expressed in this material are those of the authors and do not necessarily reflect the views of the funding agencies.
The authors thank Ankit Shah and Jessica Hodgins for their insightful feedback.


\section*{Contribution Statement}
All authors contributed equally. Listed in alphabetical order.


\bibliographystyle{named}
{\footnotesize\bibliography{references}}

\end{document}